\documentclass[10pt,twocolumn,letterpaper]{article}

\usepackage{3dv}
\usepackage{times}
\usepackage{epsfig}
\usepackage{graphicx}
\usepackage{amsmath}
\usepackage{amssymb}
\usepackage{bm}
\usepackage{gensymb}
\usepackage{makecell}
\usepackage{siunitx}
\usepackage{booktabs}
\usepackage{multirow}
\usepackage[table,xcdraw]{xcolor}
\usepackage{subfig}
\usepackage{mathrsfs}
\usepackage{threeparttable}
\usepackage{float}
\usepackage{adjustbox}
\usepackage{balance}
\usepackage{microtype}
\usepackage{listings}
\definecolor{codegreen}{rgb}{0,0.6,0}
\definecolor{codegray}{rgb}{0.5,0.5,0.5}
\definecolor{codepurple}{rgb}{0.58,0,0.82}
\definecolor{backcolour}{rgb}{0.95,0.95,0.92}
\lstdefinestyle{mystyle}{
    captionpos=t, %
    backgroundcolor=\color{backcolour},   
    commentstyle=\color{codegreen},
    keywordstyle=\color{magenta},
    numberstyle=\tiny\color{codegray},
    stringstyle=\color{codepurple},
    basicstyle=\ttfamily\footnotesize,
    breakatwhitespace=false,         
    breaklines=true,                   
    keepspaces=true,                 
    numbers=left,                    
    numbersep=5pt,                  
    showspaces=false,                
    showstringspaces=false,
    showtabs=false,                  
    tabsize=1,
}

\usepackage{enumitem}
\setenumerate[1]{itemsep=0pt,partopsep=0pt,parsep=\parskip,topsep=2.5pt}
\setitemize[1]{itemsep=1pt,partopsep=0pt,parsep=\parskip,topsep=1.5pt}
\setdescription{itemsep=0pt,partopsep=0pt,parsep=\parskip,topsep=2.5pt}

\definecolor{bg_Red}{HTML}{F6CAC9}
\definecolor{bg_Green}{HTML}{B6FAA9}

\usepackage[hyphens]{url}
\makeatletter
\g@addto@macro{\UrlBreaks}{\UrlOrds\do\r\do\u\do\b\do\i}
\makeatother

\definecolor{eccvblue}{rgb}{0.12,0.49,0.85}
\definecolor{eccvgreen}{HTML}{3A7F32}
\usepackage[pagebackref,breaklinks,colorlinks,citecolor=eccvblue,urlcolor=eccvgreen,linkcolor=eccvgreen]{hyperref}
\usepackage{cleveref}
\crefname{figure}{figure}{figures}
\crefname{table}{table}{tables}

\threedvfinalcopy %

\def\threedvPaperID{149} %

\newcommand{\datasetname}{DurLAR}

\begin{document}

\title{{\datasetname}: A High-Fidelity 128-Channel LiDAR Dataset with Panoramic Ambient and Reflectivity Imagery for Multi-Modal Autonomous Driving Applications}

\author{Li Li$^1$ \quad Khalid N. Ismail$^1$ \quad  Hubert P. H. Shum$^1$ \quad Toby P. Breckon$^{1,2}$\\
	Department of \{Computer Science$^{1}$ $|$ Engineering$^{2}$\}, Durham University, UK\\
	{\tt\small \{li.li4,\ khalid.n.ismail,\ hubert.shum,\ toby.breckon\}@durham.ac.uk}}

\maketitle

\thispagestyle{empty}

\begin{abstract}

We present {\datasetname}, a high-fidelity 128-channel 3D LiDAR dataset with panoramic ambient (near infrared) and reflectivity imagery, as well as a sample benchmark task using depth estimation for autonomous driving applications. Our driving platform is equipped with a high resolution 128 channel LiDAR, a 2MPix stereo camera, a lux meter and a GNSS/INS system. Ambient and reflectivity images are made available along with the LiDAR point clouds to facilitate multi-modal use of concurrent ambient and reflectivity scene information. Leveraging {\datasetname}, with a resolution exceeding that of prior benchmarks, we consider the task of monocular depth estimation and use this increased availability of higher resolution, yet sparse ground truth scene depth information to propose a novel joint supervised/self-supervised loss formulation. We compare performance over both our new {\datasetname} dataset, the established KITTI benchmark and the Cityscapes dataset. 
Our evaluation shows our joint use supervised and self-supervised loss terms, enabled via the superior ground truth resolution and availability within {\datasetname} improves the quantitative and qualitative performance of leading contemporary monocular depth estimation approaches ($RMSE=3.639$, $Sq Rel=0.936$).%
\end{abstract}
\vspace{-10pt}

\vspace{-5pt}
\section{Introduction}

\begin{figure}[htbp]
	\centering
	\vspace{20pt}
	\includegraphics[width=0.48\textwidth]{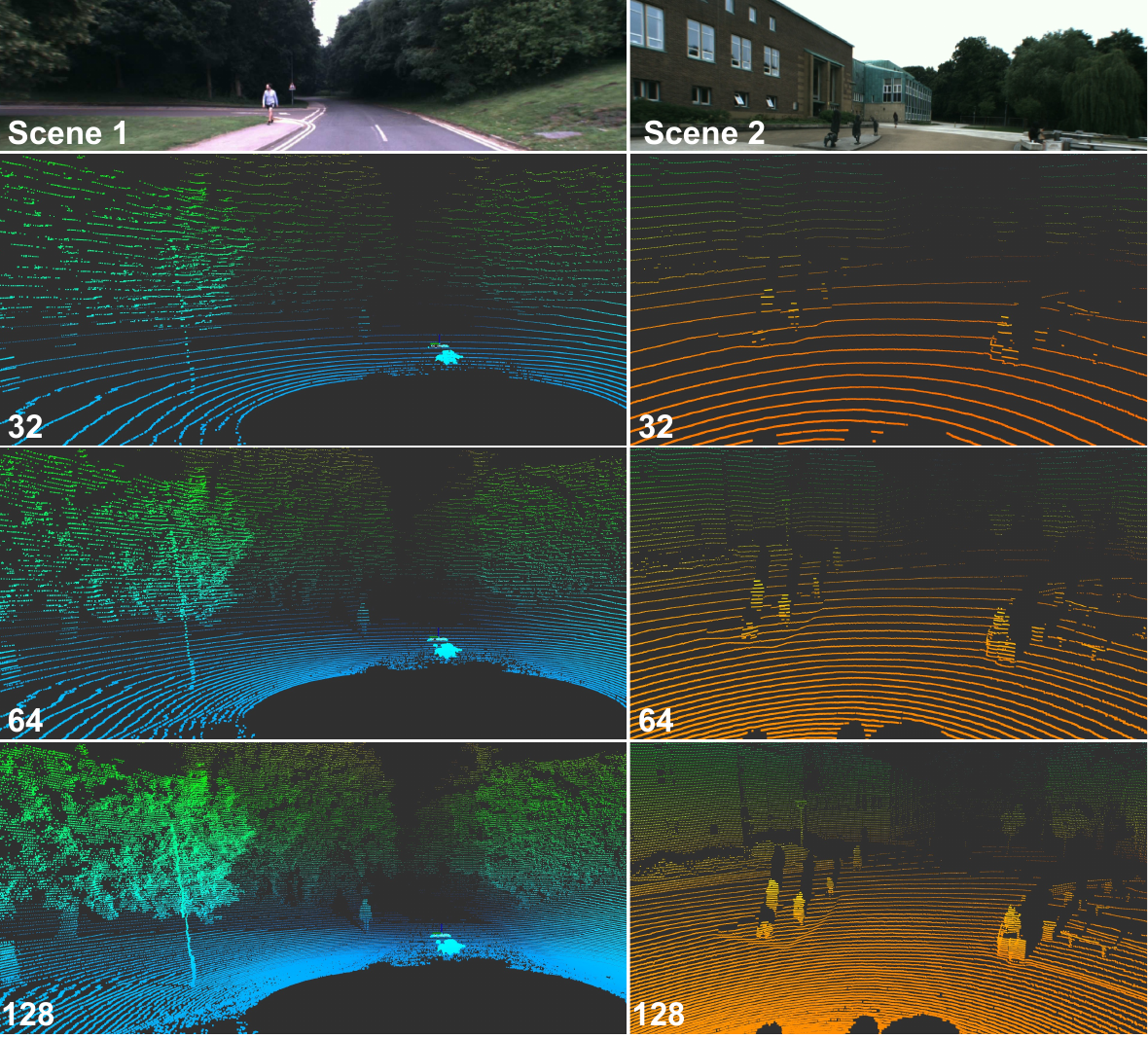}
	\caption{LiDAR point clouds from two exemplar scenes with differing vertical LiDAR resolution (top to bottom: colour RGB images, [32 $\xrightarrow{}$ 64 $\xrightarrow{}$ 128] LiDAR channels).}
	\label{fig:chCompare}
\end{figure}

\noindent LiDAR (Light Detection and Ranging) is one of the core perception technologies enabling future self-driving vehicles and advanced driver assistance systems (ADAS). Multiple datasets featuring LiDAR have been proposed to evaluate semantic in geometric scene understanding tasks such as semantic segmentation~\cite{Geiger2013a,li2023less,sun2024empirical,zhu2021cylindrical}, depth estimation~\cite{gated2depth2019}, object detection~\cite{360LiDARTracking_ICRA_2019,Teichman2011,quadros2013sydney}, visual odometry~\cite{Geiger2013a}, optical flow~\cite{Geiger2013a} and tracking~\cite{Chen_2018_CVPR,lyft2019,lyft2019prediction,nuscenes2019,RobotCarDatasetIJRR}. Based on this existing dataset provision, various architectures have been proposed for LiDAR based scene understanding in this domain~\cite{Bhat2020,Bijelic2020,monodepth2,wang2018anytime,div2020wstereo,Fu_2018_CVPR,DBLP:journals/corr/abs-1812-11941,bian2019unsupervised}. %
Moreover, benchmarks and evaluation metrics have emerged to facilitate the comparison of varies techniques and datasets~\cite{Geiger2012CVPR,Uhrig2017THREEDV,PixelAccurateDepthBenchmark2019,behley2019semantickitti,RCDRTKArXiv}. 

In these datasets, LiDAR range data corresponding to the colour image of the environment is provided as the ground-truth depth information. Such ground truth can be relatively sparse compared to the sampling of the corresponding colour camera imagery — typically as low as 16 to 64 channels of depth (see~\Cref{fig:chCompare}, \textit{e.g}, 16-64 horizontal scanlines of depth information, spanning 360 degrees from the vehicle over a 50-200~\si{m} range). Here, the terminology \textit{channel} refers to the vertical resolution of the LiDAR scanner, and has a one-to-one correspondence to the laser beam as it is referred to in some studies. With this in mind, current datasets and their associated metric-driven benchmarks are significantly limited when compared to the contemporary availability of high-resolution LiDAR data as we pursue in this paper.  %

\begin{figure}[htbp]
	\centering
	\includegraphics[width=0.45\textwidth]{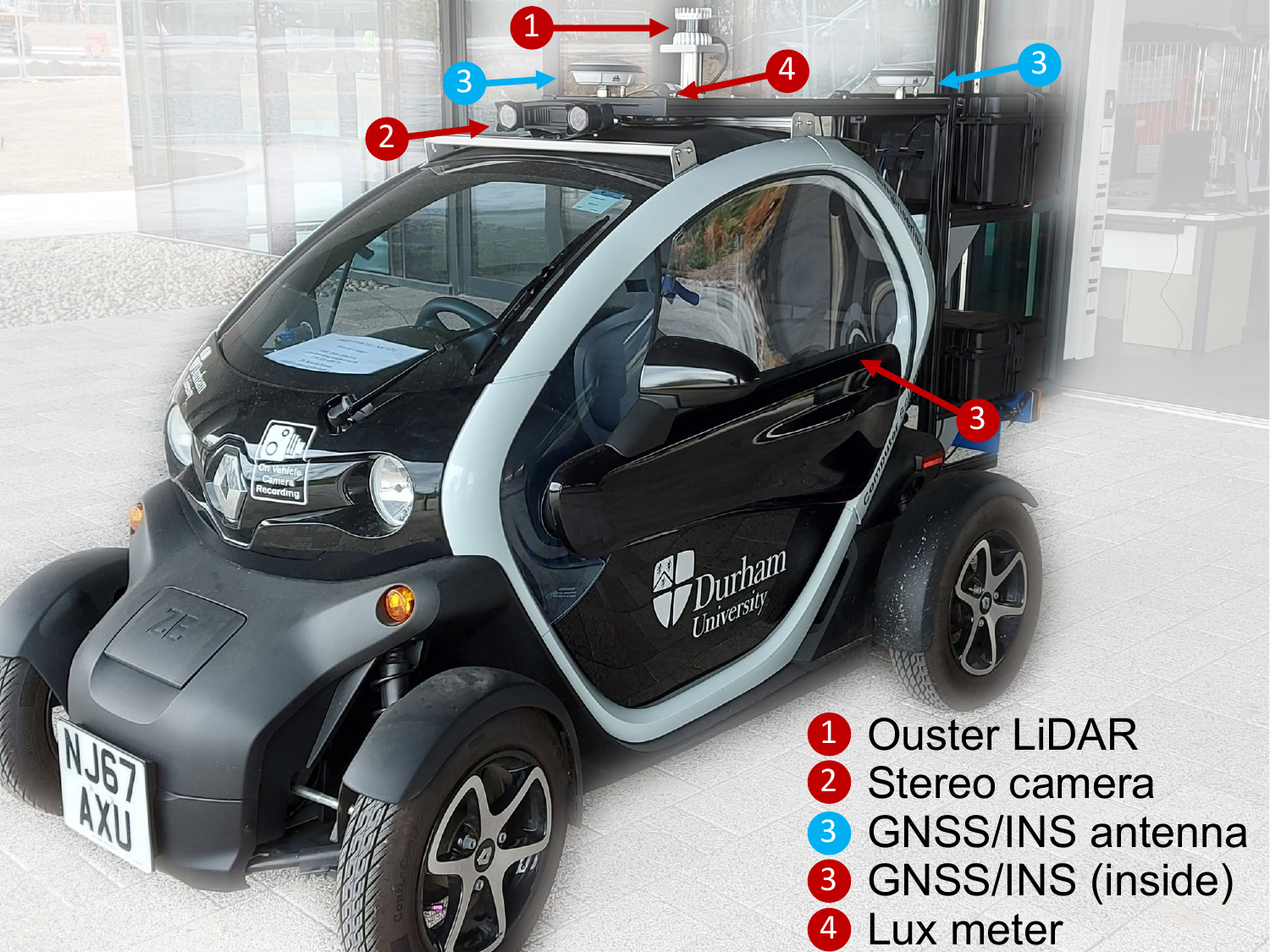}
	\caption{\textbf{Test vehicle} (Renault Twizy): equipped with a long range stereo camera, a LiDAR, a lux meter and a combined GNSS/INS inertial navigation system.}
	\label{fig:twizy}
	\vspace{-0.23cm}
\end{figure}

By contrast, we propose a large-scale high-fidelity LiDAR dataset\footnote{Online access for the dataset, \href{https://github.com/l1997i/DurLAR}{https://github.com/l1997i/DurLAR}.} based on the use of a 128 channel LiDAR unit mounted on our Renault Twizy test vehicle (\Cref{fig:twizy}). Compared to existing LiDAR datasets in this field (\Cref{tab:comparison}), including the seminal KITTI dataset~\cite{Geiger2013a,Geiger2012,Menze2015CVPR}, our dataset has the following novel features:

\begin{itemize}

    \item \textbf{High vertical resolution LiDAR}, which offers both superior spatial depth resolution (\Cref{fig:chCompare}) and additionally co-registered 360\degree ambient and reflectivity imagery that is concurrently captured via the LiDAR laser return itself.

    \item \textbf{Additional synchronised sensors} including a high resolution forward-facing stereo imagery (2MPix), a high fidelity GNSS/INS and a lux meter. 

    \item \textbf{Route repetition} such that the dataset uses the same set of driving routes under varying environmental conditions, such as overcast, rainy weather, seasonal variations and varying times of day -  hence facilitating evaluation under different weather and illumination conditions.

\end{itemize}

Subsequently, our dataset is presented as a KITTI-compatible offering such that the data formats used can be parsed using both our {\datasetname} development kit and the official KITTI tools (in addition to third party KITTI tools).

In order to illustrate the advantages and potential applications of this proposed benchmark dataset, we adopt monocular depth estimation as a sample task for comparison. We thus evaluate the relative performance of contemporary monocular depth estimation architectures~\cite{monodepth2,watson-2019-depth-hints,watson2021temporal}, by leveraging the higher resolution LiDAR capability within {\datasetname} to facilitate more effective use of depth supervision, for which we propose a novel joint supervised/self-supervised loss formulation (\Cref{sec:expriments}). 

More broadly, the illumination-independent sensing capabilities of high-resolution 3D LiDAR additionally enable the evaluation of a range of driving tasks~\cite{Thrun2006,Chen_2018_CVPR} under varying environmental conditions spanning both extreme weather and illumination changes using our dataset. %

Our main contributions are summarised as follows:
\begin{itemize}

	\item a novel large-scale dataset comprising contemporary high-fidelity 3D LiDAR (128 channels), stereo/ambient/reflectivity imagery, GNSS/INS and environmental illumination information under repeated route, variable environment conditions (in the \textit{de facto} KITTI dataset format). The first autonomous driving task dataset to additionally comprise usable ambience and reflectivity LiDAR obtained imagery (360\degree, $2048 \times 128$ resolution).
	
	\item an exemplar monocular depth estimation benchmark to compare the performance of supervised/self-supervised variants of three leading approaches~\cite{zhou2017unsupervised,monodepth2,watson2021temporal} when trained and evaluated on low resolution (KITTI~\cite{Geiger2013a}), high resolution ({\datasetname}) ground truth LiDAR depth data, or our novel KITTI/DurLAR dataset partition, with the observation that increased resolution and availability enables superior monocular depth estimation performance via the use of our joint supervised/self-supervised loss formulation (\Cref{tab:depthResults}, \Cref{tab:cross-dataset}, \Cref{fig:depth_results}).
\end{itemize}

{
	\begin{table*}[]
		\centering
		{ \small
		\begin{tabular}{@{}ccccp{1.5cm}<{\centering}p{1.3cm}<{\centering}p{2.55cm}<{\centering}@{}}
		\toprule
		Dataset                                       & Resolution   & Range/m      & Diversity        & Image          & \# Frames                                                   & Other sensors \\ \midrule
		DENSE~\cite{gated2depth2019}                  & 64           & 120          & E/W/T            & I              & 1M                                                       & D/M/F/T/B     \\
		H3D~\cite{360LiDARTracking_ICRA_2019}         & 64           & 120          & E                & I              & 28k                                                      & G/M           \\
		KITTI $\mid$ SemanticKITTI $\mid$ KITTI-360~\cite{Geiger2013a,behley2019semantickitti,Liao2021ARXIV}                      & 64           & 120          & E                & I              & 93k$\mid$93k$\mid$320k                                                      & N/S/G/M/B     \\
		LiVi-Set~\cite{Chen_2018_CVPR}                & 32           & 100          & E                & I              & 10k                                                      &               \\
		Lyft Level 5~\cite{lyft2019,lyft2019prediction}       & 64           & 200          & E/W/T            & I              & 170k                                                     & D/B           \\
		nuScenes~\cite{nuscenes2019}                  & 32           & 100          & E/W/T            & I              & 1M                                                       & M/D/B         \\
		Oxford RobotCar~\cite{RobotCarDatasetIJRR}    & 4$^a$        & 50           & E/W/T            & I              & 3M$^b$           & N/S/G/M/B     \\
		Stanford Track Collection~\cite{Teichman2011} & 64           & 120          & E                & I              & 14k                                                      & M             \\
		Sydney Urban Objects~\cite{quadros2013sydney} & 64           & 120          & E                & I              & 0.6k$^c$ &               \\
		\textbf{{\datasetname} (ours)}                  & \textbf{128} & \textbf{120} & \textbf{E/W/T/L} & \textbf{I/A/R} & \textbf{100k}                                              & \textbf{U/N/S/G/M/B}   \\ \bottomrule
		\end{tabular}%
		}
		\caption{Existing public LiDAR datasets for autonomous driving tasks detailing vertical resolution (\# channels), diversity in terms of environments (E), times of day (T), weather conditions (W), same route of repeated locations (L) and also the type of LiDAR images made available in addition to range information as: intensity (I), ambient (A), reflectivity (R). Other sensors refer to radar (D), lux meter (U), GNSS supporting more than 2 constellations (N), INS (S), GPS (G), IMU (M), FIR camera (F), Near infrared camera (T) and stereo camera (B). $^a$ the number of planes. SICK LD-MRS LiDAR has 4 planes, and SICK LMS-151 LiDAR has 1 plane. $^b$ the number of LD-MRS LiDAR frames. $^c$ the number of individual scans of objects.}
		\label{tab:comparison}%
		\vspace{-0.25cm}
	\end{table*}
}

\vspace{-7pt}
\section{Related Work}
\vspace{-5pt}

\noindent We consider prior work in two related topic areas: autonomous driving datasets (\Cref{subsec:rw_datasets}) and monocular depth estimation (\Cref{subsec:rw_depth_estim}).

\vspace{-3pt}
\subsection{Autonomous Driving Datasets}
\vspace{-3pt}
\label{subsec:rw_datasets}

\noindent There are multiple autonomous driving task datasets that provide 3D LiDAR data for outdoor environments (\Cref{tab:comparison}).

\textbf{High vertical resolution LiDAR} is not present in existing datasets (see~\Cref{tab:comparison}). The vertical resolution of LiVi-Set~\cite{Chen_2018_CVPR} and nuScenes~\cite{nuscenes2019} is 32 channels. Similarly the Stanford Track Collection~\cite{Teichman2011}, KITTI~\cite{Geiger2013a}, Sydney Urban Objects~\cite{quadros2013sydney}, DENSE~\cite{gated2depth2019}, H3D~\cite{360LiDARTracking_ICRA_2019}, SemanticKITTI~\cite{behley2019semantickitti}, Lyft Level 5~\cite{lyft2019,lyft2019prediction} and KITTI-360~\cite{Liao2021ARXIV} is 64 channels. In contrast, our proposed dataset has a higher vertical resolution of 128 channels, which can capture a significantly higher level of detail of environment objects (\Cref{fig:chCompare}). %

\textbf{Rolling shutter effect} is common among analogue spinning LiDAR, \textit{e.g.}, Velodyne scanners, which are widely used in most of the existing datasets~\cite{Geiger2013a,quadros2013sydney,Chen_2018_CVPR,nuscenes2019,gated2depth2019,360LiDARTracking_ICRA_2019,behley2019semantickitti,lyft2019,Liao2021ARXIV}. Instead, the Ouster LiDAR we use is a multi-beam flash LiDAR~\cite{ouster1}, meaning all 128 channels are shot simultaneously, avoiding this distortion effect.

\textbf{In adverse weather}, LiDAR fails~\cite{Qian_2021_CVPR} (\eg fog), since opaque particles will distort light and reduce visibility significantly, whilst it can produce fine-grained point clouds with rich information and a considerable measurement range in clear weather conditions. To handle this, some datasets have radar~\cite{Bijelic2020,lyft2019,nuscenes2019} installed, despite the much lower resolution than LiDAR. The proposed dataset publishes the ambient (near infrared) and reflectivity images besides the LiDAR point clouds (see~\Cref{tab:comparison}), which has extreme low-light sensitivity and are robust within poor illumination conditions and adverse weather.

\textbf{Data diversity} within any dataset helps the generation of more universal trained models that can operate successfully under a variety of scenarios. Some related work considers the diversity in their dataset curation~\cite{RobotCarDatasetIJRR,gated2depth2019,lyft2019,nuscenes2019}, but fail to collect data under diverse conditions over the same driving route (see~\Cref{tab:comparison}), \eg, traffic level, times of day, weathers, \textit{etc}. The proposed dataset has a wide range of data diversity via collection over the same repeated route under varying conditions.

\textbf{Ground truth depth} is not present in some seminal autonomous driving datasets, \eg, Stanford Track Collection~\cite{Teichman2011}, Sydney Urban Objects~\cite{quadros2013sydney}, Cityscapes~\cite{Cordts_2016_CVPR}, Oxford RobotCar~\cite{RobotCarDatasetIJRR}, LiVi-Set~\cite{Chen_2018_CVPR}, nuScenes~\cite{nuscenes2019} and H3D~\cite{360LiDARTracking_ICRA_2019}. Due to this limitation, they can only be applied for unsupervised and semi-supervised depth estimation methods~\cite{godard2017unsupervised,yin2018geonet}. In view of this, our proposed dataset contains ground truth depth at a higher resolution than all previous datasets (\Cref{tab:comparison}), which is applicable for both supervised and semi-supervised depth estimation tasks.

\vspace{-3pt}
\subsection{Monocular Depth Estimation}
\vspace{-3pt}
\label{subsec:rw_depth_estim}

\noindent Monocular depth estimation aims at recovering a dense depth map for each pixel using a single RGB image as input.

\textbf{Self-supervised methods} harness the monocular RGB image sequences~\cite{zhou2017unsupervised,monodepth2,abarghouei19segment-wise,abarghouei19multi-task,watson2021temporal}, stereo pairs~\cite{garg2016unsupervised,xie2016deep3d,godard2017unsupervised,watson-2019-depth-hints,Xian_2020_CVPR} or synthetic data~\cite{abarghouei18monocular,KHAN2021479} for training. Subsequently, multi-frame architectures were introduced~\cite{wang2020self,xie2020video,patil2020don,wang2019recurrent,cs2018depthnet,zhang2019exploiting,watson2021temporal}, which leverages the temporal information at test time, to improve the quality of the predicted depth. The same losses used during training can be applied to test frames to update the weights. However, additional calculations for multiple forward and backward process on a set of test frames are required which incur additional computation.

Other work concentrates on multi-view stereo (MVS), which operates on unordered image sets~\cite{mayer2016large,liang2018learning,liu2019neural,hou2019multi,khot2019learning,dai2019mvs2,wu2019spatial,wimbauer2021monorec,watson2021temporal}. Not requiring the ground truth depth and camera poses during training, self-supervised MVS methods~\cite{liu2019neural,hou2019multi,khot2019learning,dai2019mvs2,wu2019spatial,wimbauer2021monorec,watson2021temporal} leverage cost volumes to process sequences of frames at test time. Compared with the base method of MVS, these methods can predict the depth using images captured by moving cameras and do not need camera poses during training time.

\textbf{Supervised methods} utilise ground truth depth from depth sensors, \eg, LiDAR~\cite{kuznietsov2017semi,he2018wearable,fu2018deep,abarghouei19multi-task,Eldesokey_2020_CVPR} and RGB-D cameras~\cite{eigen2014depth,eigen2015predicting}, to improve the supervision feedback during learning. As with many areas of contemporary computer vision, CNN based architectures~\cite{eigen2014depth,eigen2015predicting,ummenhofer2017demon} generally offer state-of-the-art performance. Thereafter, residual-learning-based methods~\cite{he2016deep,laina2016deeper,zhang2018joint} are proposed to learn the transform relation between colour images and their corresponding maps, therefore leveraging deeper architectures than previous works with higher resultant accuracy. However, such methods are limited both by ground truth dataset availability and the fidelity (resolution) of the ground truth depth information provided.

Overall, one of key challenges within contemporary autonomous driving task evaluation is the lack of high fidelity (vertical resolution) depth datasets in order to facilitate effective evaluation of geometric scene understanding tasks, such as monocular depth estimation. Here, based on the provision of our {\datasetname} dataset (\Cref{sec:dataset}), we consider the impact of abundant high-resolution ground truth depth data on three state-of-the-art contemporary monocular depth estimation architectures (MonoDepth2~\cite{monodepth2}, Depth-hints~\cite{watson-2019-depth-hints}, ManyDepth~\cite{watson2021temporal}) through the use of our novel joint supervised/semi-supervised loss formulation (\Cref{sec:expriments}).

\vspace{-5pt}
\section{The {\datasetname} Dataset}
\vspace{-5pt}
\label{sec:dataset}

\noindent Compared to existing autonomous driving task datasets (\Cref{tab:comparison}),  {\datasetname} has the following novel features:

\begin{itemize}
	\item \textbf{High vertical resolution LiDAR} with 128 channels, which is twice that of any existing datasets (\Cref{tab:comparison}), full 360\degree depth, range accuracy to $\pm$2~\si{cm} at 20-50\si{m}.
	\item \textbf{Ambient illumination (near infrared) and reflectivity panoramic imagery} are made available in the Mono16 format ($2048 \times 128$ resolution), with this being only dataset to make this provision (\Cref{tab:comparison}).
	\item \textbf{No rolling shutter effect}, as our flash LiDAR captures all 128 channels simultaneously.
	\item \textbf{Ambient illumination data} is recorded via an onboard lux meter, which is again not available in previous datasets (\Cref{tab:comparison}).
	\item \textbf{High-fidelity GNSS/INS} available via an onboard OxTS navigation unit operating at 100~\si{Hz} and receiving position and timing data from multiple GNSS constellations in addition to GPS. %
	\item \textbf{KITTI data format} adopted as the \textit{de facto} dataset format such that it can be parsed using both the {\datasetname} development kit and existing KITTI-compatible tools.
	\item \textbf{Diversity over repeated locations} such that the dataset has been collected under diverse environmental and weather conditions over the same driving route with additional variations in the time of day relative to environmental conditions (\eg traffic, pedestrian occurrence, ambient illumination, see~\Cref{tab:comparison}).
\end{itemize}

\vspace{-3pt}
\subsection{Sensor Setup}
\vspace{-3pt}
\label{subsec:sensors}

\noindent The dataset is collected using a Renault Twizy vehicle (\Cref{fig:twizy}) equipped with the following sensor configuration (as illustrated in~\Cref{fig:placement}):

\begin{figure}[tbp]
	\centering
	\includegraphics[width=0.5\textwidth]{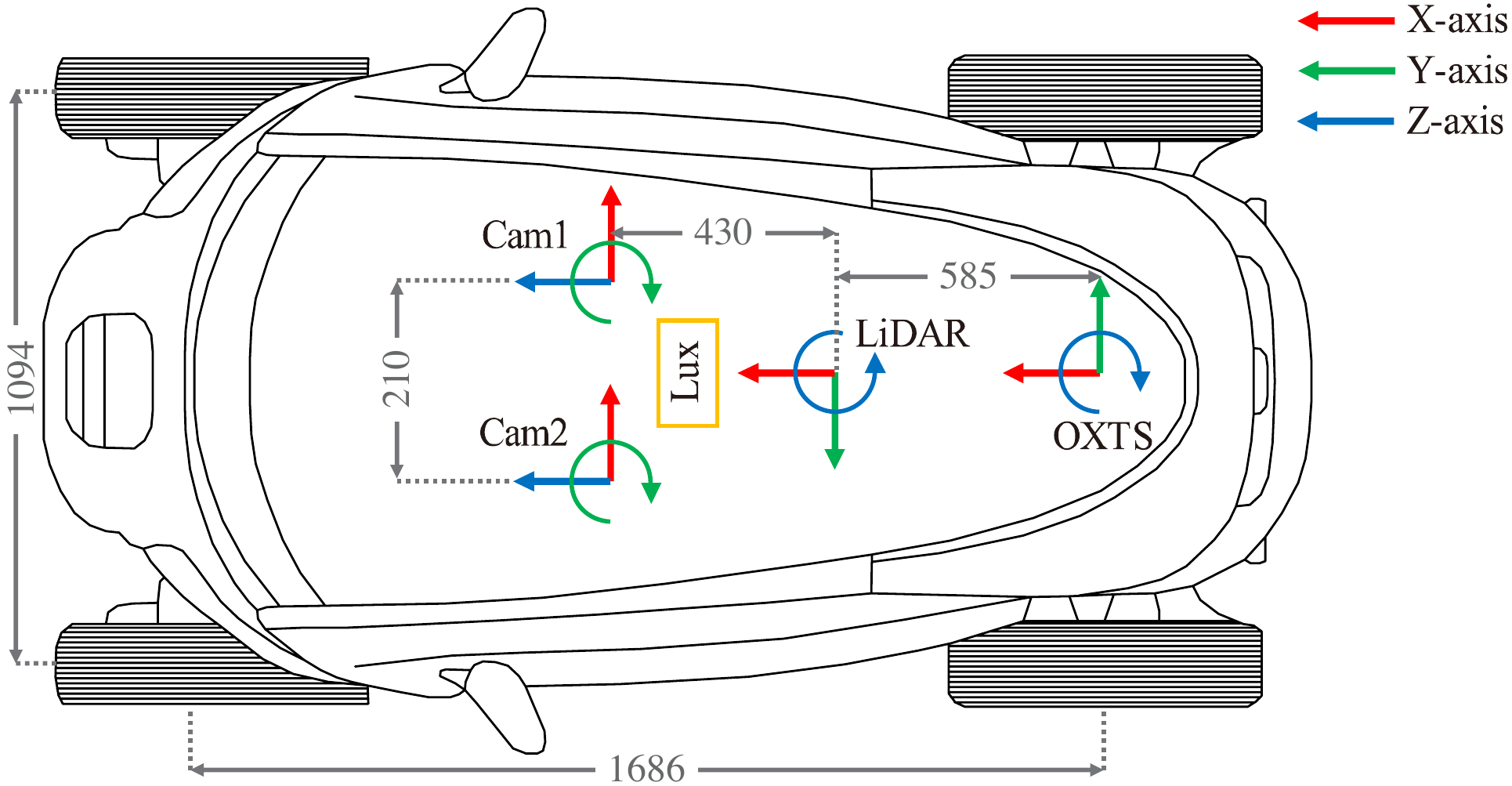}
	\caption{\textbf{Sensor placements}, top view. All coordinate axes follow the right-hand rule (sizes in~\si{mm}).}
	\label{fig:placement}
	\vspace{-0.45cm}
\end{figure}

\begin{itemize}

    \item \textbf{LiDAR}: Ouster OS1-128 LiDAR sensor with 128 channels vertical resolution, 865~\si{nm} laser wavelength, 100 m @ \textgreater 90\% detection probability and 120 m @ \textgreater 50\% detection probability (100 klx sunlight, 80\% Lambertian reflectivity, 2048 @ 10 Hz rotation rate mode), 0.3~\si{cm} range resolution, 360\degree horizontal FOV and 45\degree (+22.5\degree to -22.5\degree) vertical FOV, mounted height $\sim 1.62$~\si{m}.

    \item \textbf{Stereo Camera}: Carnegie Robotics MultiSense S21 stereo camera with grayscale, colour, and IR enhanced imagers, 0.4~\si{m} minimum range, 2048 $\times$ 1088 @ 2MP resolution, up to 30 FPS frame rate and 115\degree $\times$ 68\degree FOV, 21 cm baseline, factory calibrated, mounted height $\sim 1.42$ ~\si{m}.

    \item \textbf{GNSS/INS}: OxTS RT3000v3 global navigation satellite and inertial navigation system, with 0.03\degree pitch/roll accuracy, 0.1-1.5~\si{m} position accuracy, 0.15\degree slip angle accuracy, 250~\si{Hz} maximum data output rate, supporting positioning from GPS, GLONASS, BeiDou, Galileo, PPP and SBAS constellations. 

    \item \textbf{Lux Meter}: Yocto Light V3, a USB ambient light sensor (lux meter), measuring ambient light up to 100,000 lux, hence indirectly representing the conditions of the external environment via ambient illumination conditions.
\end{itemize}

\begin{figure}[bp]
	\centering
	\includegraphics[width=0.45\textwidth]{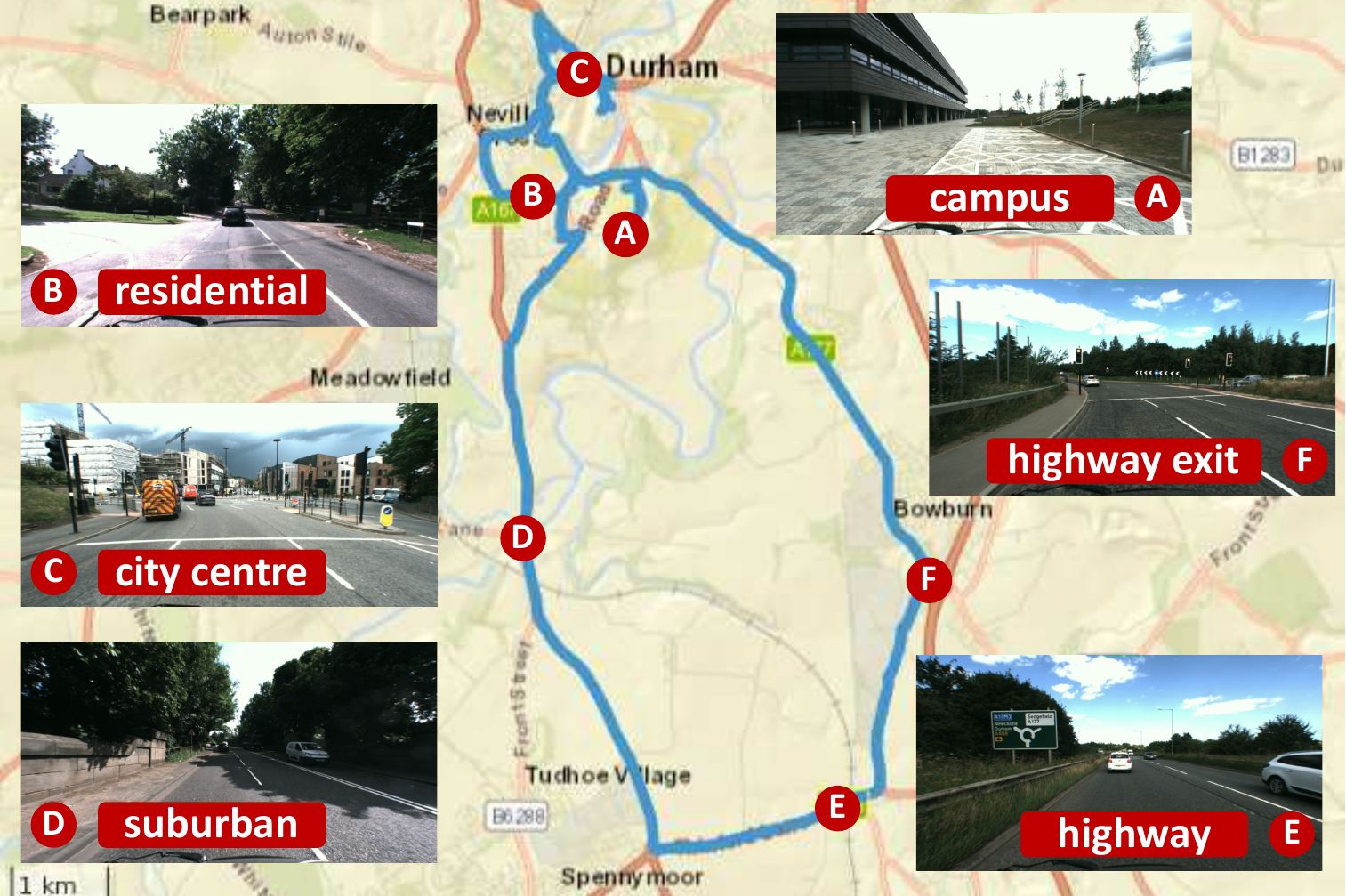}
	\caption{\textbf{The route} (blue curves) used for dataset collection showing a variety of driving environments.}
	\label{fig:gps_fix}
\end{figure}
\vspace{-0.0cm}

\vspace{-3pt}
\subsection{Data Collection and Description}
\vspace{-3pt}
\begin{figure*}[t]
	\centering
	\includegraphics[width=.95\textwidth]{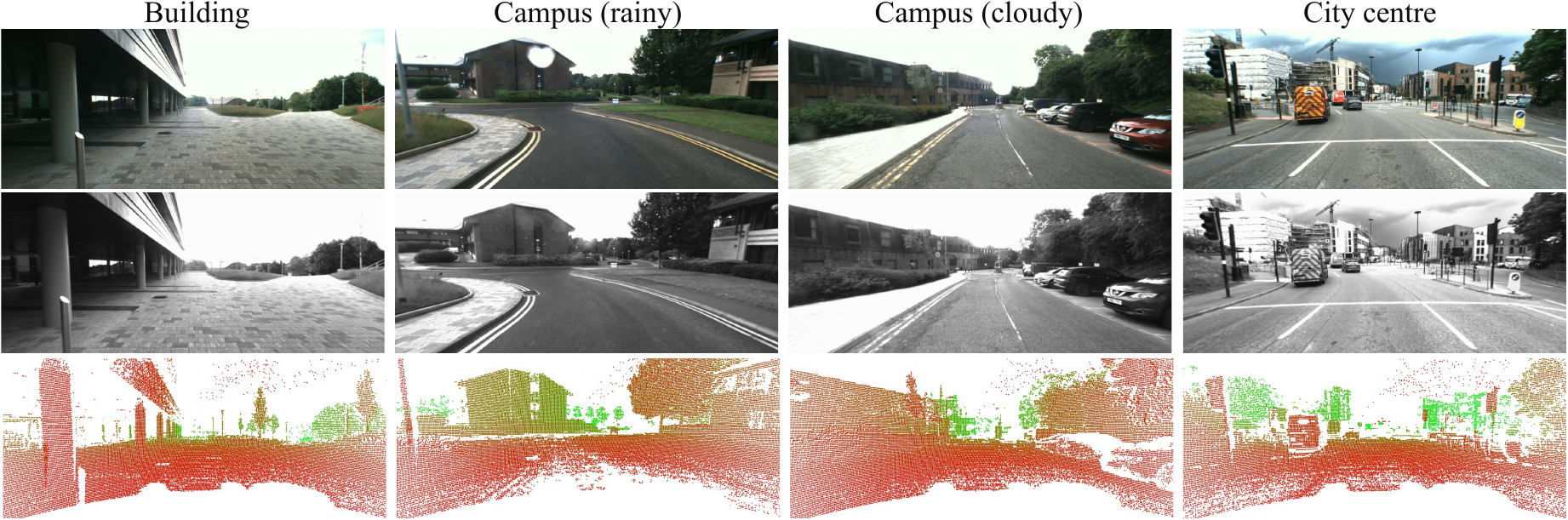}
	\caption{Examples from {\datasetname} which demonstrate the diversity in our dataset. From top to bottom, RGB left camera images (top), grayscale right camera images (centre) and LiDAR point cloud (bottom). The point cloud is projected onto the 2D image plane using the LiDAR-to-left-camera external calibration, and the colour varies with the distance from the LiDAR (near$:=$red \space\space $\to$ \space\space far$:=$green).}
	\label{fig:images_examples}
	\vspace{-0.35cm}
\end{figure*}

\noindent To ensure the dataset has diverse weather and varying density of pedestrian and traffic occurrences, we collect the data over a variety of conditions. These includes different types of environments, times of day, weather and repeated locations along the test route with data collected for the key time periods and environments shown in~\Cref{tab:data_time_periods}. As shown in~\Cref{fig:gps_fix} and~\Cref{fig:images_examples}, our dataset mainly contains suburban, highway, city centre and campus areas.

\begin{table}[tbp]
	\centering
	\small
	\begin{tabular}{@{}ccccc@{}}
	\toprule
	\multicolumn{1}{c}{} & \begin{tabular}[c]{@{}c@{}}Avg. Speed\end{tabular} & Day.      & Peak times & Night \\ \midrule
	City        & 20.4 km/h  & [3] $|$ [3]  & [3] $|$ [3]     & [2] $|$ [3]    \\
	Campus      & 26.4 km/h  & [1] $|$ [1]  & [1] $|$ [2]     & [1] $|$ [1]    \\
	Residential & 31.2 km/h  & [1] $|$ [2]  & [2] $|$ [2]     & [1] $|$ [1]    \\
	Suburb      & 43.6 km/h  & [1] $|$ [1]  & [1] $|$ [1]     & [1] $|$ [1]    \\ \bottomrule
	\end{tabular}
	\caption{Key time periods and environmental conditions. The value is expressed in the form of [traffic density] $|$ [population density], using a qualitative scale of [3 - high, 2 - normal, 1 - low].}
	\label{tab:data_time_periods}
	\vspace{-0.35cm}
\end{table}

All the data is provided in the \textit{de facto} KITTI data formats, with the exception of the ambient light data (lux) which is not provided by KITTI and is hence published in a simple plain text format with aligned timestamp.

\vspace{-2pt}
\subsection{Ambient and Reflectivity Panoramic Imagery}
\vspace{-1.5pt}
\begin{figure*}[t]
	\centering
	\includegraphics[width=0.99\textwidth]{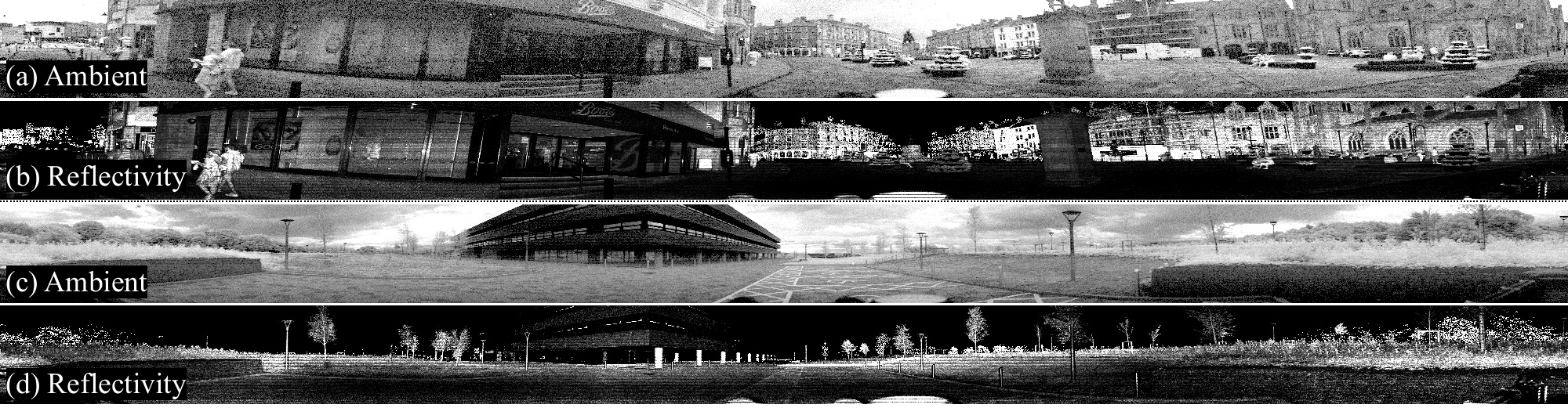}
	\caption{Example of ambient (near infrared) and reflectivity panoramic images.}
	\label{fig:amb_and_reflec}
\end{figure*}

\noindent The proposed {\datasetname} dataset is the first autonomous driving task  dataset to additionally provide high-resolution ambient and reflectivity panoramic 360-degree imagery. The ambient imagery can be captured even in low light conditions (near infrared, ~800-2500~\si{nm}), while the reflectivity imagery pertains to the material property of the scene object and its reflectivity of the 850~\si{nm} LiDAR signal in use (Ouster OS1-128). These characteristics, combined with a superior vertical resolution when compared to other datasets, enable these images to offer great benefit when dealing with unfavourable illumination conditions and coherent scene object identification.

\textbf{Ambient images} offer day/night scene visibility in the near-infrared spectrum. The photon counting ASIC (Application Specific Integrated Circuit) of our sensor has particularly strong illumination sensitivity, so that the ambient images can be captured even in low light conditions. This is extremely practical in designing techniques that are specifically appropriate for adverse illumination conditions, such as nocturnal and adverse weather conditions.

\textbf{Reflectivity images} contain information indicative of the material properties of the object itself and offer good consistency across illumination conditions and range. However, the Ouster OS1-128 LiDAR does not collect the true reflectivity data directly due to sensor limitations. Instead, an estimation of the reflectivity data is used to calculate the reflectivity images from the LiDAR intensity and range data. LiDAR intensity is the return signal strength of the laser pulse that recorded the range reading. According to the inverse square law (\Cref{equ:ier2}) for Lambertian objects in the far field, the intensity per unit area varies inversely proportional to the square of the distance~\cite{ouster2}, 
\vspace{-0.1cm}
\begin{equation}\label{equ:ier2}
	I = \frac{S}{4\pi r^2},
\end{equation}
\vspace{-0.0cm}
where $I$ is the intensity, $r$ is the range (namely the distance of the object to the sensor) and $S$ is the source strength.

The calculation of reflectivity assumes that it is proportional to the source strength, which is also proportional to the product of intensity and the square of the range,
\vspace{-0.1cm}
\begin{equation}\label{equ:reflec_cal}
	\text{Reflectivity} \propto S \propto I r^2 .
\end{equation}
\vspace{-0.0cm}
Exemplar ambient (near infrared) and reflectivity panoramic imagery is shown in~\Cref{fig:amb_and_reflec}. In~\Cref{fig:amb_and_reflec} (a) and (c), clouds and shadows of objects can be distinguished (expressed as shades of grayscale). These pictures are very close to the images of grayscale or RGB camera. In~\Cref{fig:amb_and_reflec} (b) and (d), the reflectivity of the same object or material will remain constant regardless of the distance to the sensor, weather, light illumination and other conditions, since reflectivity is the intrinsic property of the object itself. The pillars of the building (\Cref{fig:amb_and_reflec} (d)) have almost the same reflectivity (\textit{i.e.} the same white colour in the figure) regardless of their distance to the LiDAR sensor.

\subsection{Calibration and Synchronisation}
\label{sec:calib_and_sync}

\begin{figure}[b]
	\centering
	\includegraphics[width=0.45\textwidth]{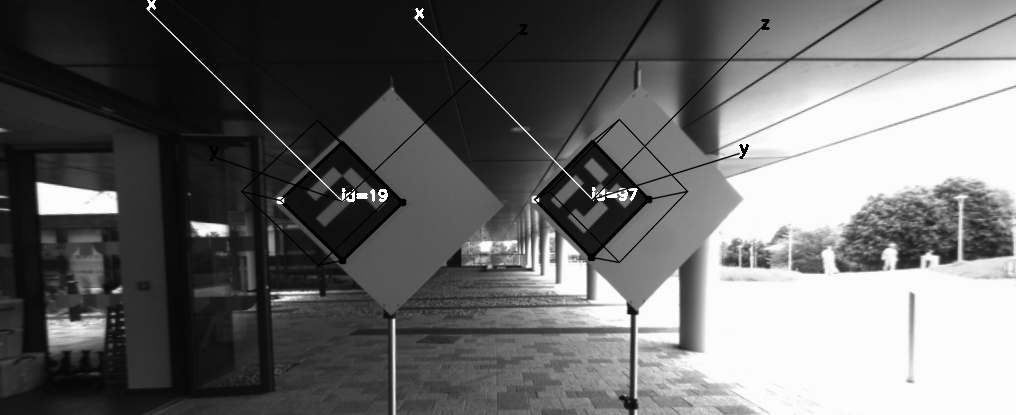}
	\caption{Camera to LiDAR custom calibration pattern with extrinsic parameter estimation overlay shown.}
	\label{fig:ext_calibration}
\end{figure}

\begin{figure}[tb]
	\centering
	\includegraphics[width=0.45\textwidth]{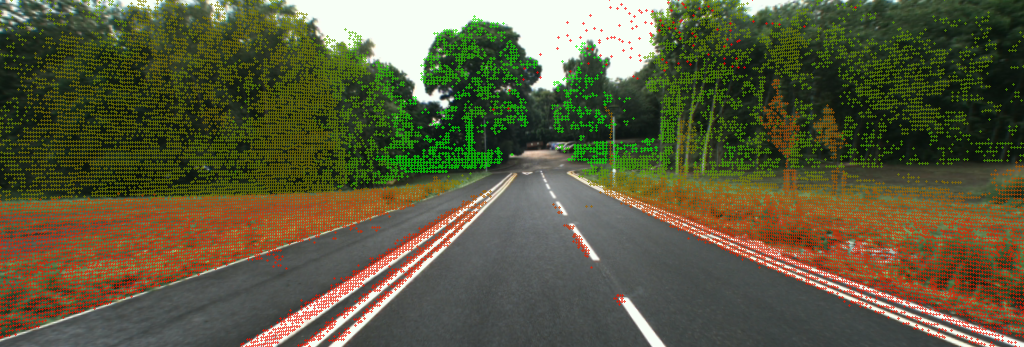}
	\caption{Illustrative LiDAR 3D point cloud overlay onto the right stereo image (colour) using the calibration obtained.}
	\label{fig:calibration_proj}
\end{figure}

\noindent LiDAR-to-camera calibration is performed using ~\cite{2017arXiv170509785D,beltran2021automatic}. With the custom calibration pattern shown in~\Cref{fig:ext_calibration}, the calibration is composed of two stages (refer to Appendix~\ref{sec:calibration_details}). Firstly, a pair of two ArUco markers are detected from the left frame of the stereo camera such that the transformation matrix $[R|t]$, containing rotation $R$ and translation $t$ parameters, between the camera and the centre of the ArUco marker can be calculated (as shown in the overlays of~\Cref{fig:calibration_proj}). Secondly, the edges of the orientated calibration boards are identified in the corresponding LiDAR data frame projection by orientated edge detection. Finally, the optimal rigid transformation between the LiDAR and the camera is found using RANSAC based optimisation~\cite{2017arXiv170509785D}. 

Stereo camera calibration is based on the manufacturer factory instructions for intrinsic and extrinsic settings. Calibration of the GNSS/INS is performed using the manufacturers recommended approach. The GNSS/INS with respect to the LiDAR is registered following~\cite{FITZGIBBON20031145}.

All sensor synchronisation is performed at a rate of 10~\si{Hz}, using Robot Operating System (ROS, version Noetic) timestamps operating over a Gigabit Ethernet backbone to a common host (Intel Core i5-6300U, 16 GB RAM).

\vspace{-0.35cm}

\section{Monocular Depth Estimation}
\label{sec:expriments}

\begin{table*}[htbp]
\centering
\resizebox{\textwidth}{!}{%
\begin{tabular}{ccccccccccc}
\toprule
Dataset &
Method &
+S &
W $\times$ H &
\cellcolor[HTML]{F6CAC9}Abs Rel &
\cellcolor[HTML]{F6CAC9}Sq Rel &
\cellcolor[HTML]{F6CAC9}RMSE &
\cellcolor[HTML]{F6CAC9}RMSE log &
\cellcolor[HTML]{B6FAA9}$\delta < 1.25$ &
\cellcolor[HTML]{B6FAA9}$\delta < 1.25^2$ &
\cellcolor[HTML]{B6FAA9}$\delta < 1.25^3$ \\
\midrule
    & ManyDepth (MR)~\cite{watson2021temporal} & $\times$ & 640 $\times$ 192  & 0.098 & 0.770 & 4.459 & 0.176 & 0.900 & 0.965 & \textbf{0.983} \\
	\multirow{-2}{*}{KITTI~\cite{Geiger2013a}} & ManyDepth (HR)~\cite{watson2021temporal} & $\times$ & 1024 $\times$ 320 & \textbf{0.093} & \textbf{0.715} & \textbf{4.245} & \textbf{0.172} & \textbf{0.909} & \textbf{0.966} & \textbf{0.983} \\
\midrule
Cityscapes~\cite{Cordts_2016_CVPR} & ManyDepth~\cite{watson2021temporal} & $\times$ & 416 $\times$ 128  & 0.114 & 1.193 & 6.223 & 0.170 & 0.875 & 0.967 & 0.989 \\
\midrule
	& Depth-hints~\cite{watson-2019-depth-hints} & $\times$ & 640 $\times$ 192  &   0.122  &   1.070  &   4.148  &   0.211  &   0.870  &   0.946  &   0.972  \\  %
	& Depth-hints~\cite{watson-2019-depth-hints} & \checkmark & 640 $\times$ 192  &   0.121  &   1.109  &   4.121  &   0.210  &   0.874  &   0.946  &   0.972  \\ %
	& MonoDepth2~\cite{monodepth2} & $\times$ & 640 $\times$ 192  &   0.111  &   1.114  &   4.002  &   0.187  &   0.895  &   0.960  &   0.981  \\ %
	& MonoDepth2~\cite{monodepth2} & \checkmark & 640 $\times$ 192  &   \underline{0.108}  &   \underline{1.010}  &   \underline{3.804}  &   0.185  &   0.898  &   0.963  &   0.982  \\ %
    & ManyDepth (MR)~\cite{watson2021temporal} & $\times$ & 640 $\times$ 192  &   0.115  &   1.227  &   4.116  &   0.186  &   0.892  &   0.962  &   0.982  \\ %
	& ManyDepth (MR)~\cite{watson2021temporal} & \checkmark & 640 $\times$ 192  &   0.109  &   \textbf{0.936}  &   3.711  &   \underline{0.176}  &   0.895  &   0.964  &   \underline{0.984}  \\ %
    & ManyDepth (HR)~\cite{watson2021temporal} & $\times$ & 1024 $\times$ 320 &   0.109  &   1.111  &   3.875  &   0.177  &   \underline{0.901}  &   \underline{0.966}  &   \underline{0.984}  \\	%
	\multirow{-8}{*}{{\datasetname}} & ManyDepth (HR)~\cite{watson2021temporal} & \checkmark & 1024 $\times$ 320 &   \textbf{0.104}  &   \textbf{0.936}  &   \textbf{3.639}  &   \textbf{0.171}  &   \textbf{0.906}  &   \textbf{0.969}  &   \textbf{0.986}	\\ %
\bottomrule
\end{tabular}%
} 
\caption{Performance comparison over the KITTI Eigen split~\cite{Geiger2012}, Cityscapes~\cite{Cordts_2016_CVPR} (self-supervised only) and {\datasetname} datasets (+S, joint supervised/self-supervised (\checkmark) \textit{v.s.} self-supervised ($\times$)). MR and HR stand for medium and high resolution of training models (as originally defined in~\cite{watson2021temporal}). Depth evaluation metrics are shown in the top row. \colorbox{bg_Red}{Red} refers to superior performances indicated by low values, and \colorbox{bg_Green}{green} refers to superior performance indicated by a higher value. The best results in KITTI and {\datasetname} are in \textbf{bold}; the second best in {\datasetname} are \underline{underlined}.}
\label{tab:depthResults}
\vspace{-0.35cm}
\end{table*}

\noindent Leveraging the higher vertical LiDAR resolution of our {\datasetname} dataset, we adopt monocular depth estimation as an illustrative benchmark task. 

We select ManyDepth~\cite{watson2021temporal} as a leading approach for monocular depth estimation as it offers state-of-the-art performance on the leading KITTI~\cite{Geiger2013a} and Cityscapes~\cite{Cordts_2016_CVPR} benchmarks. Whilst ManyDepth~\cite{watson2021temporal} is a self-supervised approach, here we seek to leverage the availability of high-fidelity depth within  {\datasetname} via the introduction of a secondary supervised loss term to formulate a novel supervised/self-supervised loss formulation. As a result, we can assess the impact of the availability of abundant ground truth depth at training time on the performance of this leading contemporary approach.

To these ends, we introduce the reverse Huber (Berhu) loss $\mathscr{L}_{\text {Berhu }}$~\cite{Lambert-Lacroix} as our supervised depth loss term, due to its effectiveness in smoothing and blurring depth prediction edges on object boundaries: 
\begin{equation}\label{equ:berhu_loss}
	\mathscr{L}_{\text {Berhu }}\left(d, d^{*}\right)= \begin{cases}\left|d-d^{*}\right| & \text { if }\left|d-d^{*}\right| \leq \delta,  \\ \frac{\left(d-d^{*}\right)^{2}+\delta ^{2}}{2 \delta } & \text { if }\left|d-d^{*}\right|>\delta, \end{cases}
\end{equation}
where $d$ is the predicted depth, $d^{*}$ is the ground truth depth, and $\delta$ stands for the threshold. If $\left|d-d^{*}\right| \leq \delta$, the Berhu loss is equal to $\mathscr{L}_1$; else, it acts approximately as $\mathscr{L}_2$.

We hence construct a joint supervised/semi-supervised version of ManyDepth~\cite{watson2021temporal}, adding $\mathscr{L}_{\text {Berhu }}$ into the original ManyDepth loss function, as shown in~\Cref{equ:manydepth_loss}:
\begin{equation}\label{equ:manydepth_loss}
	\mathscr{L}=(1-M) \mathscr{L}_{p}+\mathscr{L}_{\text {consistency}}+\mathscr{L}_{\text {smooth}}+\mathscr{L}_{\text {Berhu }},
\end{equation}
where $\mathscr{L}_{p}$ is the photometric reprojection error and $\mathscr{L}_{\text {smooth}}$ is the smoothness loss, from~\cite{monodepth2,watson2021temporal}. $\mathscr{L}_{\text {consistency}}$ is the consistency loss, as implemented from~\cite{watson2021temporal}.

For an extended comparison, we similarly introduce this additional supervised depth loss via this additional Berhu loss term to the contemporary MonoDepth2~\cite{monodepth2} and Depth-hints~\cite{watson-2019-depth-hints} approaches leaving the remainder of the architectures unchanged.

We specify a randomly generated data split for the {\datasetname} dataset as well, comprising 90k training frames, 5k validation frames and 5k test frames for our evaluation.

\vspace{-3pt}
\section{Evaluation Results}
\vspace{-3pt}

\noindent Training was performed with all learning parameters set as per the original works~\cite{monodepth2,watson2021temporal, watson-2019-depth-hints}, with Berhu threshold $\delta = 0.2$, on a Nvidia Tesla V100 GPU over 20 epochs.

\vspace{-3pt}
\subsection{Quantitative Evaluation}
\vspace{-3pt}

\noindent The varying performance of self-supervised depth estimation between the KITTI~\cite{Geiger2013a}, Cityscapes~\cite{Cordts_2016_CVPR} and proposed {\datasetname} dataset illustrates the varying levels of challenge and complexity afforded by variations within the datasets (\Cref{tab:depthResults}, records with $\times$ in the +S column) %

However, within our evaluation on the {\datasetname} dataset, we consistently observe superior performance (lower RMSE, higher accuracy, \textit{etc,}~\Cref{tab:depthResults}) with the use of additional depth supervision (\textit{i.e.} joint supervised/semi-supervised loss, see~\Cref{tab:depthResults} - records with \checkmark in the +S column) across all three monocular depth estimation approaches considered and show overall state-of-the-art performance on monocular depth estimation using our joint supervised/self-supervised ManyDepth variant ({\datasetname},~\Cref{tab:depthResults} - as highlighted in bold).

\vspace{-3pt}
\subsection{Qualitative Evaluation}

\begin{figure*}[t]
	\centering
	\includegraphics[width=.9\textwidth]{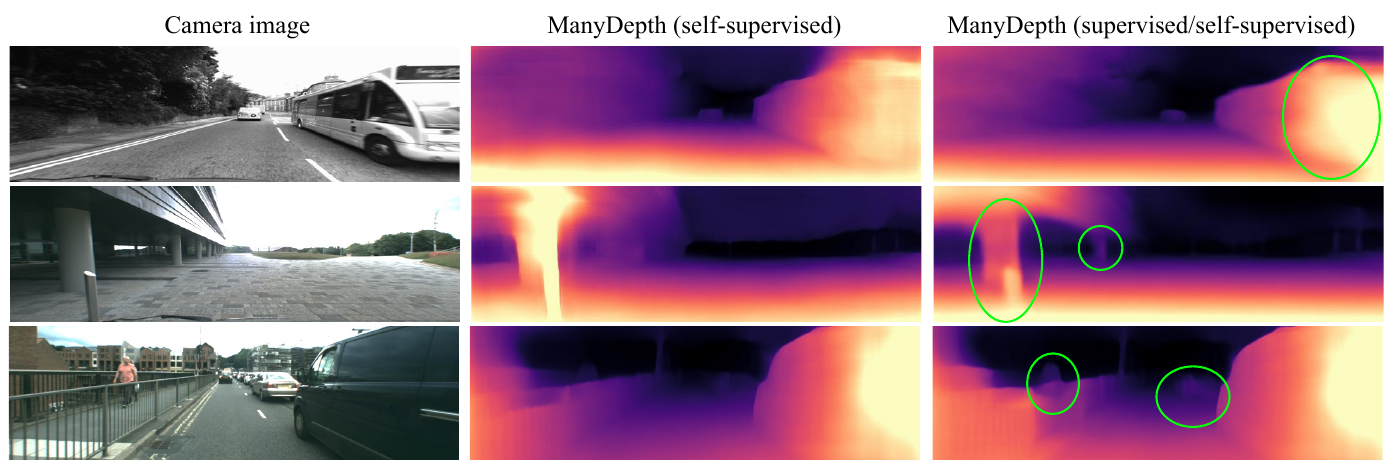}
	\caption{Comparison of monocular depth estimation results with areas of improvement highlighted with the use of depth supervision (green).}
	\label{fig:depth_results}
\end{figure*}

\noindent To qualitatively illustrate the difference between self-supervised and joint supervised/self-supervised ManyDepth with the addition of depth loss, we show exemplars highlighting areas of superior depth estimation (\Cref{fig:depth_results}).

Within these examples, we can see a clearer contour edge of the bus and resolution of the upper LED display board on the vehicle (\Cref{fig:depth_results}, top - self-supervised \textit{v.s.} supervised/self-supervised). Furthermore, we see improved depth resolution of the building (\Cref{fig:depth_results}, middle - self-supervised \textit{v.s.} supervised/self-supervised) whereby 
additional depth supervision enables the technique to correctly estimate the depth of the supporting building pillars and is even able to resolve the depth of the short stainless steel stub in the foreground. Finally, we can see improved estimation and clarity of both vehicle and pedestrian depth within a crowded urban scene (\Cref{fig:depth_results}, bottom - self-supervised \textit{v.s.} supervised/self-supervised).

Furthermore, we conduct additional comparative cross-training experiments to explore training on DurLAR, KITTI or KITTI/DurLAR combined whilst evaluating on a novel KITTI/DurLAR union split (\Cref{tab:cross-dataset}). Our KITTI/DurLAR union training/testing data split presents a challenging evaluation task that is more diverse, with 694 test frames each from KITTI and DurLAR, to measure the overall performance across both datasets. 

\begin{table}[!hb]
	\footnotesize
	\centering
	{%
	\begin{tabular}{@{}cp{0.72cm}<{\centering}p{0.72cm}<{\centering}p{0.72cm}<{\centering}p{0.72cm}<{\centering}ccc@{}}
	\toprule
	Train &
    \cellcolor[HTML]{F6CAC9}\begin{tabular}[c]{@{}c@{}} Abs\\ Rel\end{tabular} &
    \cellcolor[HTML]{F6CAC9}\begin{tabular}[c]{@{}c@{}} Sq\\ Rel\end{tabular} &
    \cellcolor[HTML]{F6CAC9} RMSE &
    \cellcolor[HTML]{F6CAC9}\begin{tabular}[c]{@{}c@{}} RMSE\\ log\end{tabular} &
    \cellcolor[HTML]{B6FAA9}$\delta_1$ &
    \cellcolor[HTML]{B6FAA9}$\delta_2$ &
    \cellcolor[HTML]{B6FAA9}$\delta_3$ \\ \midrule

	K    &   0.159  &   1.536  &   5.101  &   0.244  &   0.798  &   0.923  &   0.963  \\
	D    &   0.189  &   1.764  &   5.580  &   0.264  &   0.758  &   0.908  &   0.959  \\
	K+D  &   0.188  &   1.941  &   5.182  &   0.262  &   0.769  &   0.912  &   0.958  \\
	D+K  &   \textbf{0.151}  &   \textbf{1.123}  &   \textbf{4.744}  &   \textbf{0.233}  &   \textbf{0.805}  &   \textbf{0.927}  &   \textbf{0.967}  \\
	\bottomrule
	\end{tabular}%
	}
	\caption{\textbf{Cross-dataset tests} of ManyDepth [51] with the training configuration (K) KITTI only, (D) DurLAR only, (K+D) KITTI then fine-tuning with DurLAR, (D+K) DurLAR then fine-tuning with KITTI. $\delta_1$, $\delta_2$ and $\delta_3$ refers to $\delta< 1.25$, $\delta< 1.2^2$ and $\delta< 1.25^3$ respectively.}
	\label{tab:cross-dataset}
\end{table}

\vspace{-5.5pt}
\subsection{Ablation Study}
\vspace{-3.5pt}

\noindent Our ablation study shows the side-by-side impact of our joint supervised/unsupervised loss formulation in addition to the performance impact of high-fidelity depth (higher vertical LiDAR resolution).

\textbf{Supervised depth}: We train the ManyDepth~\cite{watson2021temporal} with and without the Berhu loss (Equation \ref{equ:berhu_loss}), such that we can compare the original self-supervised performance with that of additional depth supervision (\Cref{tab:ablation}, 128/-S \textit{v.s.} 128/+S).

\textbf{Ground truth depth resolution:} We simulate a reduction in vertical ground truth depth resolution by subsampling the depth values present by 50\% (64 channels) and 75\% (32 channels) along the vertical axis of the LiDAR ground truth projection. From~\Cref{tab:ablation}, we can see superior performance from our joint supervised/unsupervised loss formulation (128/-S \textit{v.s.} 128/+S) and from higher vertical resolution LiDAR (32/64 \textit{v.s.} 128/-S).

\begin{table}[bt]
	\footnotesize
	\centering
	{%
	\begin{tabular}{@{}cp{0.72cm}<{\centering}p{0.72cm}<{\centering}p{0.72cm}<{\centering}p{0.72cm}<{\centering}ccc@{}}
	\toprule
	vRes &
    \cellcolor[HTML]{F6CAC9}\begin{tabular}[c]{@{}c@{}} Abs\\ Rel\end{tabular} &
    \cellcolor[HTML]{F6CAC9}\begin{tabular}[c]{@{}c@{}} Sq\\ Rel\end{tabular} &
    \cellcolor[HTML]{F6CAC9} RMSE &
    \cellcolor[HTML]{F6CAC9}\begin{tabular}[c]{@{}c@{}} RMSE\\ log\end{tabular} &
    \cellcolor[HTML]{B6FAA9}$\delta_1$ &
    \cellcolor[HTML]{B6FAA9}$\delta_2$ &
    \cellcolor[HTML]{B6FAA9}$\delta_3$ \\ \midrule

	32/+S  &   0.115  &   \textbf{0.908}  &   \underline{3.677}  &   0.179  &   0.888  &   0.966  &   \underline{0.985}  \\ %
	64/+S  &   \underline{0.107}  &   \underline{0.918}  &   3.735  &   \underline{0.175}  &   0.895  &   \underline{0.967}  &   \textbf{0.986}  \\ %
	128/-S & 0.109 & 1.111 & 3.875 & 0.177 & \underline{0.901} & 0.966 & 0.984 \\
	128/+S & \textbf{0.104} & 0.936 & \textbf{3.639} & \textbf{0.171} & \textbf{0.906} & \textbf{0.969} & \textbf{0.986} \\ \bottomrule
	\end{tabular}%
	}
	\caption{\textbf{Ablation results on ManyDepth}~\cite{watson2021temporal}. vRes $:=$ the vertical resolution of LiDAR ground truth depth. $\pm$S $:=$ supervised/self-supervised (+S) and self-supervised ManyDepth (-S) for consistency with~\Cref{tab:depthResults}.}
	\label{tab:ablation}
	\vspace{-0.5cm}
\end{table}

\section{Conclusion}
\noindent In this paper, we present a high-fidelity 128-channel 3D LiDAR dataset with panoramic ambient (near infrared) and reflectivity imagery for autonomous driving applications ({\datasetname}). In addition, we present the exemplar benchmark task of depth estimation task whereby we show the impact of higher resolution LiDAR as a means to the supervised extension of leading contemporary monocular depth estimation approaches~\cite{monodepth2,watson-2019-depth-hints,watson2021temporal}. 

{\datasetname}, is a novel large-scale dataset comprising contemporary high-fidelity LiDAR, stereo/ambient/reflectivity imagery, GNSS/INS and environmental illumination information under repeated route, variable environment conditions (in the \textit{de facto} KITTI dataset format). It is the first autonomous driving task dataset to additionally comprise usable ambience and reflectivity LiDAR obtained imagery ($2048 \times 128$ resolution).

In our sample monocular depth estimation task, we show superior performance can be achieved by leveraging the high resolution LiDAR resolution afforded by {\datasetname} via the secondary introduction of an additional supervised loss term for depth. This is demonstrated across three state-of-the-art monocular depth estimation approaches~\cite{monodepth2,watson-2019-depth-hints,watson2021temporal}. We show that the recent availability of abundant high-resolution ground truth depth from sensors such as those used in {\datasetname} enable new research possibilities for supervised learning within this domain.

Further work will consider the provision of additional dataset annotation spanning object, semantic and geometric scene information. Future application utilising the ambient and reflectivity imagery will be explored. %

\vspace{-10pt}
\paragraph{\small Acknowledgements:}

\small
This work made use of the facilities of the N8 Centre of Excellence in Computationally Intensive Research (N8 CIR) provided and funded by the N8 research partnership and EPSRC (Grant No. EP/T022167/1). The Centre is co-ordinated by the Universities of Durham, Manchester and York.

\clearpage
% ┌──────────────────────────────────────────────────────────────────────────────┐
% │                  Author - Li (Luis) Li @ Durham University                   │
% │                                                                              │
% │                        Department of Computer Science                        │
% │                                                                              │
% │                                v3.0.0 - 14 Jun 2024                          │
% │                                                                              │
% │                             li.li4@durham.ac.uk                              │
% └──────────────────────────────────────────────────────────────────────────────┘

\renewcommand\appendix{\par
    \setcounter{section}{0}
    \setcounter{subsection}{0}
    \gdef\thesection{\Alph{section}}}
\appendix

\vspace{-1cm}
\twocolumn[
\begin{center}
    \huge \textbf{Appendix}
\end{center}
\vspace{1cm}
]

\begin{figure*}[h]
    \centering
    \includegraphics[width=1\linewidth]{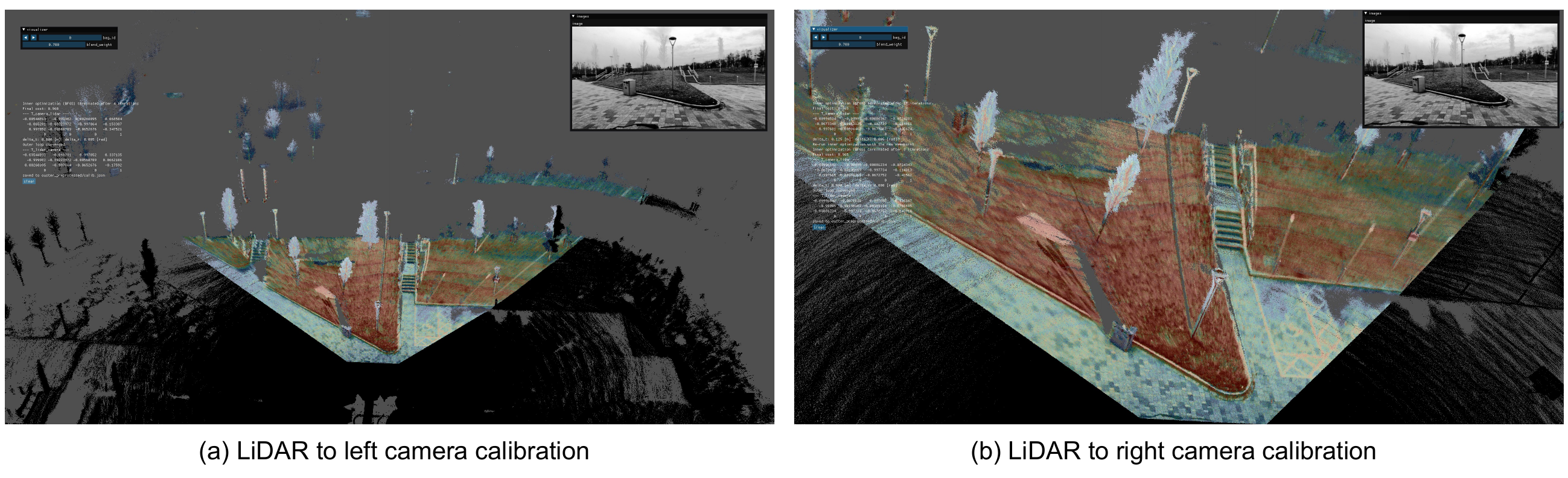}
    \caption{LiDAR to stereo camera calibration and visualisation.}
    \label{fig:lidar_to_cam}
\end{figure*}

In this appendix, we supplement additional materials to support our findings, observations, and experimental results.
Specifically, it is organised as follows:
\begin{itemize}
    \item Appendix~\ref{sec:calibration_details} supplements the updated methods on LiDAR-camera calibration which achieves better calibration results than our previously used method.
    \item Appendix~\ref{sec:access_durlar} supplements the details on getting access to our {\datasetname} dataset.
\end{itemize}

\section{LiDAR-Camera Calibration Details}
\label{sec:calibration_details}

\noindent Following the publication of the proposed {\datasetname} dataset and this paper (the original version submitted to the conference), we identify a more advanced targetless calibration method~\cite{koide2023general} that surpasses the LiDAR-camera calibration technique previously employed in Section~\ref{sec:calib_and_sync}. As shown in Figure~\ref{fig:lidar_to_cam}, by overlaying the LiDAR intensity features and the camera grayscale features with a certain level of transparency, we can see that our updated calibration results are ideal and accurate. 

\begin{figure}[h]
    \centering
    \includegraphics[width=1\linewidth]{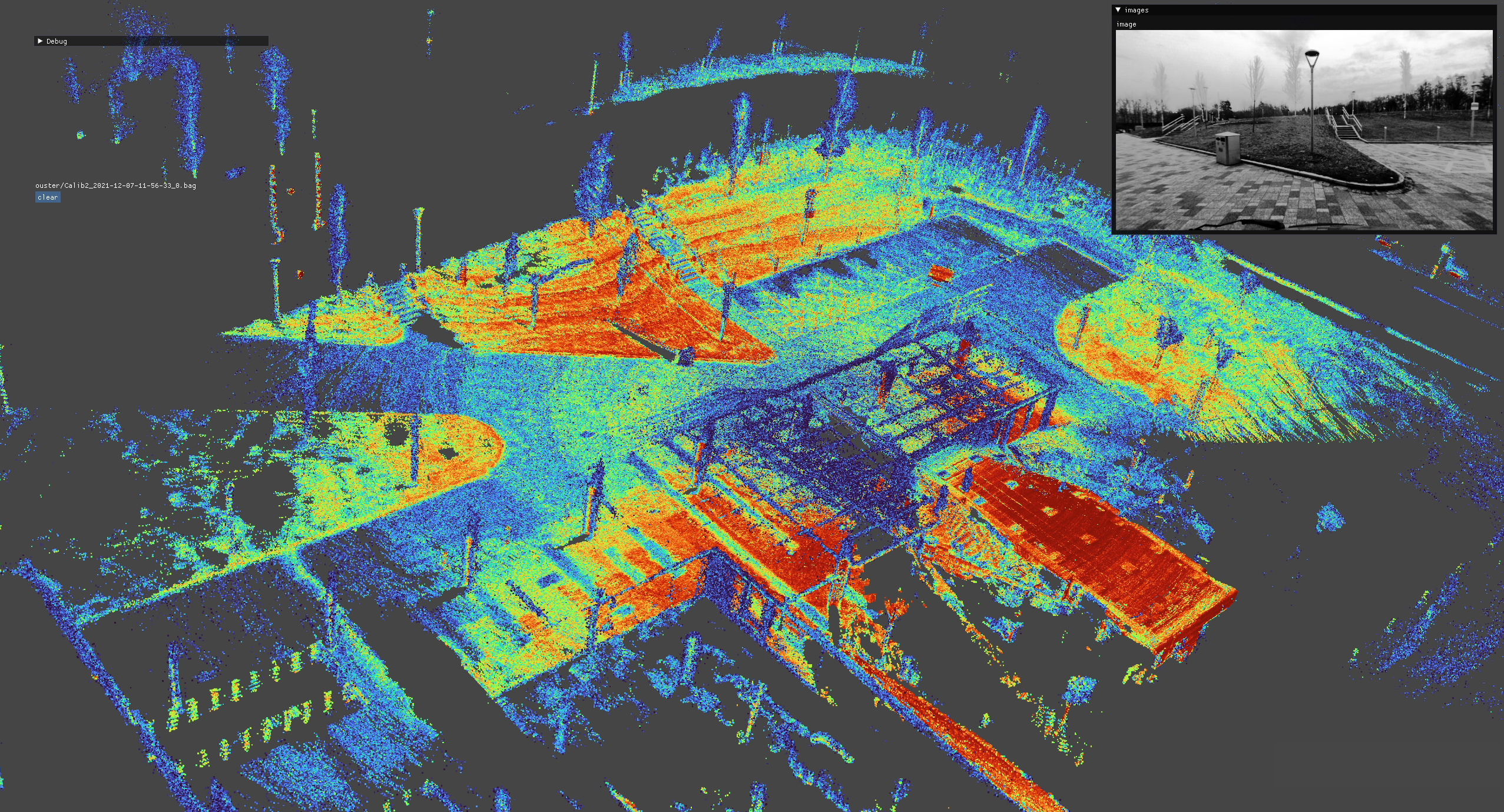}
    \caption{LiDAR frame-wise aggregation allows for the generation of a denser point cloud from continuous dynamic LiDAR frames, resulting in detailed geometrical and surface texture information.}
    \label{fig:calib_preprocessing}
\end{figure}

Given that our Ouster OS1-128 operates as a spinning LiDAR, it faces challenges associated with its sparse and repetitive scan patterns~\cite{koide2023general}, rendering the extraction of meaningful geometrical and texture information from a single scan particularly difficult. To address this, as shown in Figure~\ref{fig:calib_preprocessing}, we pre-process a continuous series of sparse point cloud frames by accumulating points while compensating for viewpoint changes and distortion~\cite{koide2023general}.

\begin{figure}[h]
    \centering
    \includegraphics[width=1\linewidth]{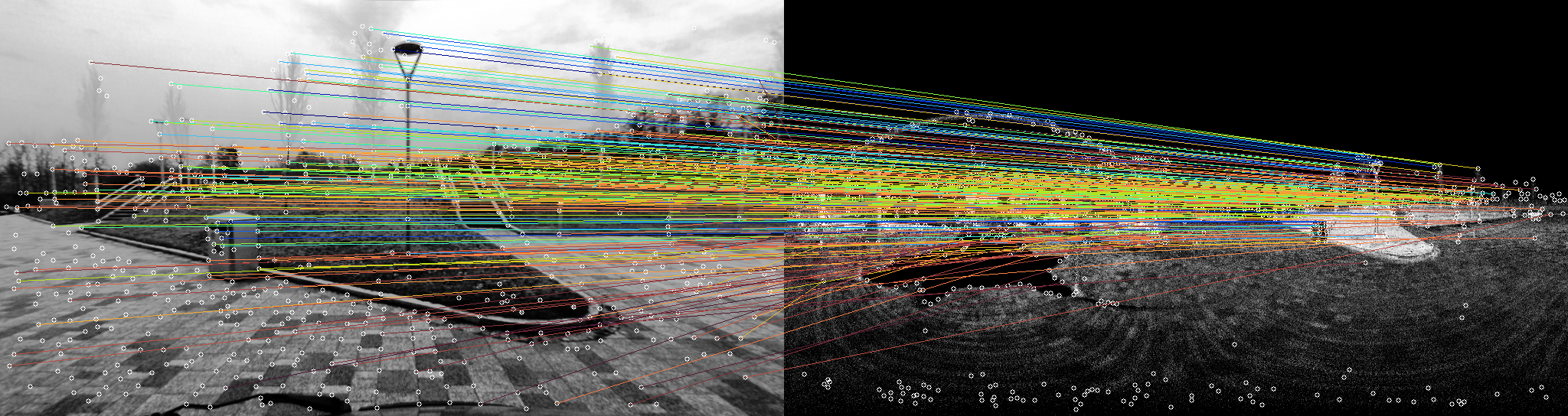}
    \caption{SuperGlue is used to identify correspondences between LiDAR points and camera images.}
    \label{fig:calib_superglue}
\end{figure}

Given the densified point cloud and camera image, we find 2D-3D correspondences using SuperGlue~\cite{sarlin2020superglue}. As shown in Figure~\ref{fig:calib_superglue}, SuperGlue identifies correspondences between LiDAR points and camera images across different modalities, even with a relative low matching threshold. The results include numerous false correspondences that must be filtered out before pose estimation (green: inliers $\to$ red: outliers).

Based on the 2D-3D correspondences, an initial estimate of the LiDAR-camera transformation is derived using RANSAC and reprojection error minimisation. Finally, the precise LiDAR-camera registration is achieved through NID~\cite{pascoe2017nid} minimisation.

We officially provide both the new and old versions of the calibration results and the original bag files for calibration, allowing users to utilise them as per their requirements.

\section{Public Access for DurLAR Dataset}
\label{sec:access_durlar}

\noindent Our DurLAR dataset is open-accessed to public, which is hosted on \href{https://collections.durham.ac.uk/collections/r2gq67jr192}{Durham Collections}. In this section, we provide details for accessing the DurLAR dataset, as well as descriptions of the data, related tools, and scripts.

\subsection{Data Structure}

\noindent In DurLAR dataset, each drive folder contains 8 topic folders for every frame,
\begin{itemize}
    \item \texttt{ambient/}: panoramic ambient imagery
    \item \texttt{reflec/}: panoramic reflectivity imagery
    \item \texttt{image\_01/}: right camera (grayscale+synced+rectified)
    \item \texttt{image\_02/}: left RGB camera (synced+rectified)
    \item \texttt{ouster\_points}: ouster LiDAR point cloud (KITTI-compatible binary format)
    \item \texttt{gps, imu, lux}: \texttt{csv} format files
\end{itemize}

\begin{figure}[htp]
    \centering
    \includegraphics[width=0.75\linewidth]{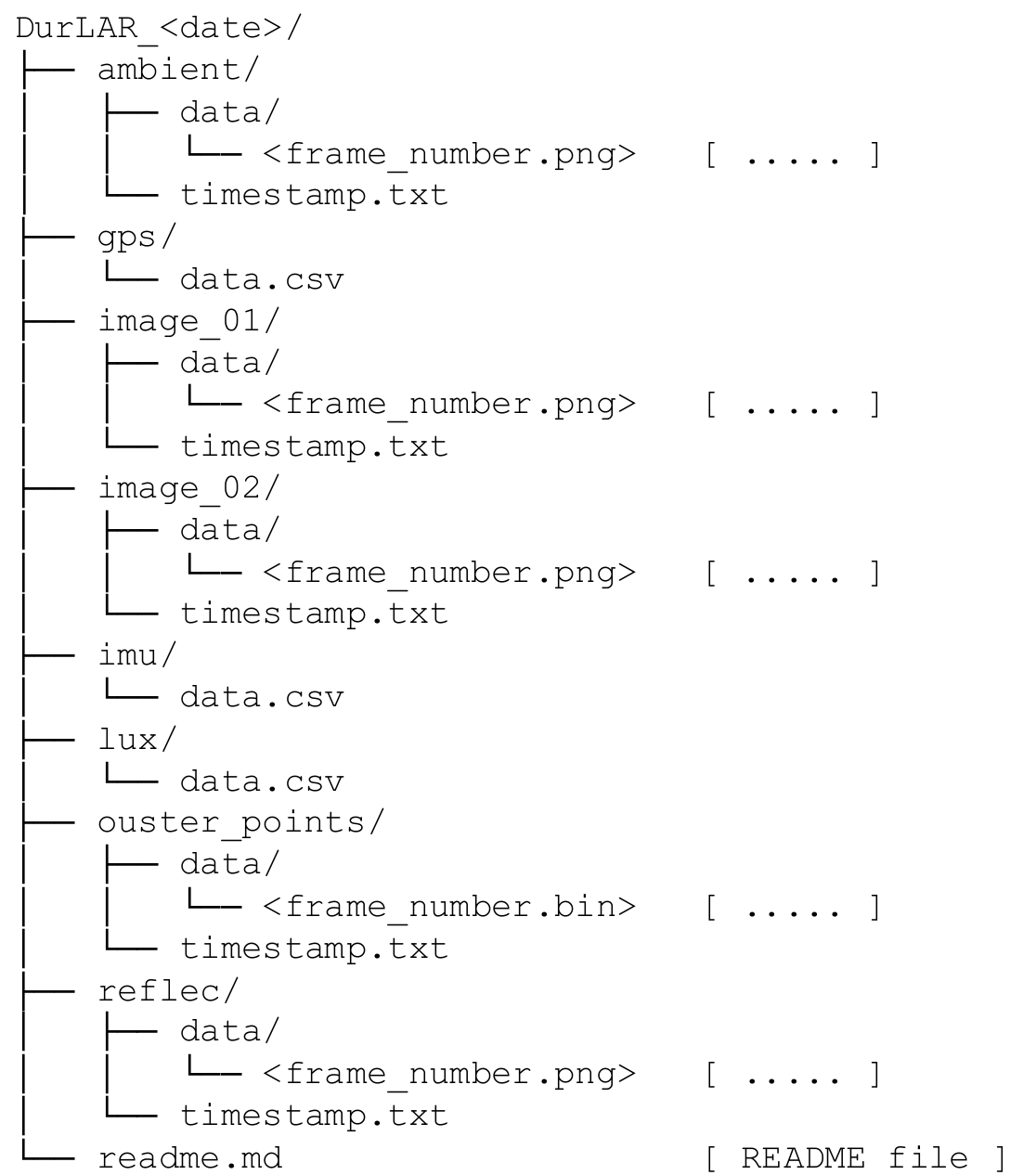}
    \caption{The folder structure of the DurLAR dataset.}
    \label{fig:durlar_folder_struct}
\end{figure}

\begin{figure}[htp]
    \centering
    \includegraphics[width=1.1\linewidth]{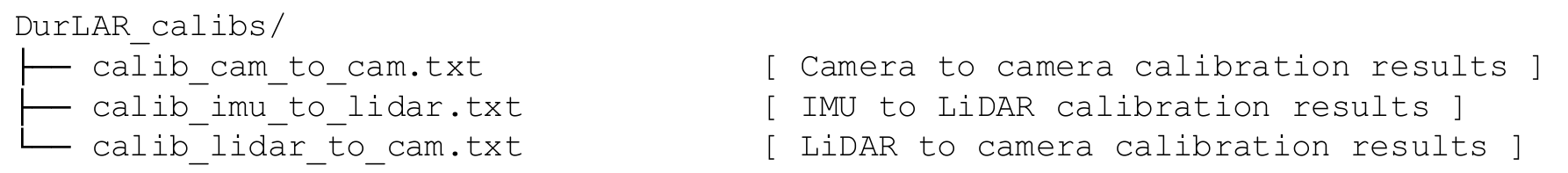}
    \caption{The folder structure of the DurLAR calibration information.}
    \label{fig:durlar_calib_struct}
\end{figure}

The folder structure of the DurLAR dataset is shown in Figure~\ref{fig:durlar_folder_struct}. The folder structure of the DurLAR calibration information (both internal and external calibration) is shown in Figure~\ref{fig:durlar_calib_struct}.

\subsection{Download the Dataset}

\noindent Access to the complete DurLAR dataset can be requested through the following link:~\href{https://forms.gle/ZjSs3PWeGjjnXmwg9}{https://forms.gle/ZjSs3PWeGjjnXmwg9}). Upon completion of the form, the download script \texttt{durlar\_download} and accompanying instructions will be automatically provided. The DurLAR dataset can then be downloaded via the command line using Terminal.

For the first use, it is highly likely that the \texttt{durlar\_download} file will need to be made executable: 
\begin{lstlisting}[language=bash]
chmod +x durlar_download
\end{lstlisting}
By default, this script downloads the exemplar dataset (600 frames, \href{https://collections.durham.ac.uk/collections/r2gq67jr192}{direct link}) for unit testing:
\begin{lstlisting}[language=bash]
./durlar_download
\end{lstlisting}
It is also possible to select and download various test drives:
\begin{lstlisting}[language=bash]
usage: ./durlar_download [dataset_sample_size] [drive]
dataset_sample_size = [ small | medium | full ]
drive = 1 ... 5
\end{lstlisting}
Given the substantial size of the DurLAR dataset, please download the complete dataset only when necessary:
\begin{lstlisting}[language=bash]
./durlar_download full 5
\end{lstlisting}

\textls[-5]{During the entire download process, your network must not encounter any issues. If there are network problems, please delete all DurLAR dataset folders and rerun the download scripts. The download script is now only support Ubuntu (tested on Ubuntu 18.04 and Ubuntu 20.04, amd64) for now. Please refer to \href{https://collections.durham.ac.uk/collections/r2gq67jr192}{Durham Collections} to download the DurLAR dataset for other OS manually.}

\subsection{Integrity Verification}
\noindent For easy verification of folder data and integrity, we provide the number of frames in each drive folder in Table~\ref{tab:verif_integr}, as well as the \href{https://collections.durham.ac.uk/collections/r2gq67jr192?utf8=%E2%9C%93&cq=MD5&sort=}{MD5 checksums} of the zip files.

\begin{table}[htp]
\caption{The number of frames in each drive folder.}
\label{tab:verif_integr}
\setlength{\tabcolsep}{8pt}
\begin{adjustbox}{width=1\linewidth}
\begin{tabular}{@{}ccccccc@{}}
\toprule
\textbf{Drive ID}      & \texttt{20210716} & \texttt{20210901} & \texttt{20211012} & \texttt{20211208} & \texttt{20211209} & \textbf{Total}  \\ \midrule
\textbf{\# of Frames} & 41993    & 23347    & 28642    & 26850    & 25079    & \textbf{145911} \\
\bottomrule
\end{tabular}
\end{adjustbox}
\end{table}

\noindent \textbf{Acknowledgement}: We thank Mr. \href{https://www.lgdv.tf.fau.de/person/richard-marcus/}{Richard Marcus} for pointing out the recent calibration method~\cite{koide2023general}.
\clearpage

\balance
{\small
\bibliographystyle{ieee}
\bibliography{DurhamCamp,bib_appendix}
}

\end{document}